\documentclass[letterpaper]{article}
\usepackage{aaai2026}
\nocopyright
\usepackage{times}
\usepackage{helvet}
\usepackage{courier}
\usepackage[hyphens]{url}
\usepackage{graphicx}
\usepackage{natbib}
\usepackage{caption}
\usepackage{booktabs}
\usepackage{amsmath}
\usepackage{amsfonts}
\usepackage{amssymb}
\usepackage{multirow}
\usepackage{microtype}
\usepackage{adjustbox}
\usepackage{xcolor}
\usepackage{colortbl}
\usepackage{needspace}
\usepackage{fancyhdr}
\usepackage{cuted}

\AtBeginDocument{\raggedbottom}
\fancypagestyle{usipreprint}{%
    \fancyhf{}%
    \fancyhead[L]{\footnotesize\normalfont Z. Zhu et al.}%
    \fancyhead[R]{\footnotesize\normalfont USI-Brain}%
    \fancyfoot[C]{\footnotesize\normalfont\thepage}%
}
\fancypagestyle{usifirst}{%
    \fancyhf{}%
    \fancyfoot[C]{\footnotesize\normalfont\thepage}%
}

\ifdefined\pdfobjcompresslevel
\fi

\definecolor{gainred}{HTML}{C62828}
\definecolor{dropgreen}{HTML}{18733C}
\newcommand{\posdelta}[1]{\textcolor{gainred}{#1}}
\newcommand{\negdelta}[1]{\textcolor{dropgreen}{#1}}
\newcommand{\gain}[1]{\posdelta{(#1)}}
\newcommand{\drop}[1]{\negdelta{(#1)}}

\definecolor{bestblue}{HTML}{CBE3F5}
\definecolor{secondblue}{HTML}{EAF2F8}
\newcommand{\bestcell}{\cellcolor{bestblue}}
\newcommand{\secondcell}{\cellcolor{secondblue}}

\title{USI-Brain: Enhancing Underwater Spatial Intelligence via Spatial Data-Driven Vision-Language Modeling}

\author{
    Zhiliang Zhu\textsuperscript{1,*,\ensuremath{\dagger}},
    Zhiqiang Zhao\textsuperscript{1,*},
    Xie'an Jiang\textsuperscript{1,*},\\
    Yingxin Lai\textsuperscript{2},
    Kangye Zheng\textsuperscript{1},
    Dawei Zhao\textsuperscript{1,\ensuremath{\ddagger}}
}
\affiliations{
    \textsuperscript{1}Harbin Engineering University\\
    \textsuperscript{2}Xiamen University\\
    \url{zhuzhiliang@hrbeu.edu.cn}
}

\begin{document}
\maketitle
\pagestyle{usipreprint}
\thispagestyle{usifirst}

\begingroup
\renewcommand{\thefootnote}{}
\makeatletter
\long\def\@makefntext#1{\noindent#1}
\makeatother
\footnotetext[0]{%
    \textsuperscript{*}Equal contribution.\\
    \textsuperscript{\ensuremath{\dagger}}Project leader.\\
    \textsuperscript{\ensuremath{\ddagger}}Corresponding author: Dawei Zhao
    (\url{daweizhao@hrbeu.edu.cn}).%
}
\endgroup

\begin{abstract}
Underwater spatial reasoning depends on object boundaries and geometric cues that are obscured by scattering, color distortion, and low visibility. Existing underwater resources capture these cues through images, annotations, and depth maps, but provide fragmented supervision for vision-language models. We present a spatial data engine that connects visual appearance with object-level geometry through enhancement, 2D and geometry-assisted question generation, and expert filtering. It produces 4.27M QA pairs across five capability groups. We use this corpus to train USI-Brain and introduce an Underwater Prior-guided Rectifier (UPR) that applies gated corrections to visual features. USI-Bench evaluates the resulting capabilities with 6,525 held-out QA pairs. On its 15 task metrics, USI-Brain and USI-Brain + UPR improve over Qwen3-VL-8B-Instruct by 5.76 and 7.63 points on average, respectively, using signed differences on 100-point scales. Ablations show that geometry-assisted supervision improves depth and ordering by 4.8 and 6.1 points with UPR disabled, while UPR adds 18.0 points in scene description under matched training data. Both variants use raw images at evaluation. Across NautData, UWBench, and MarineEval, USI-Brain achieves the best or joint-best score on 28 of 32 metrics. Model weights and USI-Bench will be released at \url{https://github.com/SKY-Zhu-91/USI-Brain}.
\end{abstract}

\begin{figure*}[!tp]
  \centering
  \includegraphics[width=\textwidth,height=0.54\textheight,keepaspectratio]{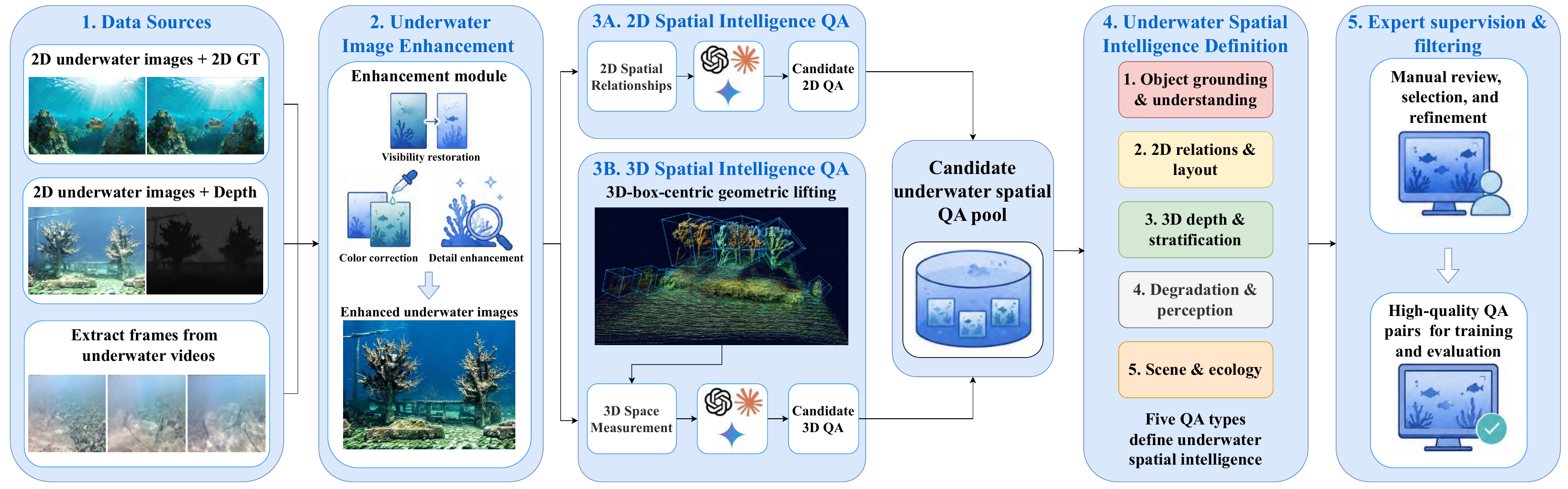}
  \caption{Overview of the underwater spatial data engine. The engine enhances heterogeneous inputs, generates 2D and geometry-assisted 3D QA, organizes them into five capability groups, and applies expert filtering.}
  \label{fig:pipeline-overview}
\end{figure*}

\section{Introduction}
\label{sec:introduction}

Marine ecological monitoring, offshore inspection, and autonomous ocean operations require models to locate objects, compare their depths, and interpret the surrounding scene \citep{yuan2023marineuvp,nauert2023inspection,wynn2014auvs,huy2023underwaterperception,zhang2023auvnavigation}. Identifying a structure is one part of this process; determining its position relative to nearby organisms, its visibility, and its spatial extent requires evidence about both appearance and geometry. We study these capabilities as \emph{underwater spatial intelligence}.

Underwater imaging degrades the cues needed for such reasoning. Wavelength-dependent attenuation alters color, while backscatter, blur, and low illumination obscure boundaries and texture \citep{akkaynak2019seathru,li2020uieb,gu2023underwatercomputationalimaging}. Image enhancement addresses these appearance changes. Questions about depth, ordering, and occlusion additionally require geometric relationships to be represented in the supervision \citep{cai2025underwaterdepth}. This motivates a data-construction process that combines enhancement with object-linked geometry.

The required evidence is distributed across existing resources. NAUTILUS~\citep{xu2025nautilus} provides large-scale underwater multimodal data; MarineInst~\citep{zheng2024marineinst}, CoralSCOP~\citep{zheng2024coralmask}, and SynTIDE~\citep{lin2025syntide} contribute instance annotations, segmentation, and depth-related supervision. UWBench~\citep{zhang2025uwbench}, CoralVQA~\citep{han2025coralvqa}, UQ-Bench~\citep{cao2026uqbench}, and MarineEval~\citep{wong2026marineeval} evaluate different aspects of underwater understanding. Tables~\ref{tab:train-resource-comparison} and~\ref{tab:eval-comparison} show the resulting differences in capability coverage. A common QA representation can bring these resources together while retaining the correspondence between an object, its image region, and its geometric cues.

Structured spatial supervision has improved general vision-language models in SpatialVLM~\citep{chen2024spatialvlm}, Visual Spatial Tuning~\citep{yang2025vst}, and OpenSpatial~\citep{liu2026openspatial}. We apply this principle to underwater data through a \emph{spatial data engine}. Hybrid Underwater Enhancement (HUE) prepares images for annotation. A 2D branch generates questions about appearance and image-plane relations; a geometry-assisted branch connects image regions with depth samples and reconstructed object boxes to generate depth, ordering, occlusion, and stratification questions. Capability-based organization and expert review yield 4.27M QA pairs spanning five groups.

We train USI-Brain from Qwen3-VL-8B-Instruct~\citep{bai2025qwen3vl} on this corpus. To adapt visual representations to degraded inputs, we introduce an Underwater Prior-guided Rectifier (UPR) that predicts additive and multiplicative feature corrections and combines them through a residual gate. USI-Brain + UPR retains both raw and corrected visual tokens. HUE shapes the training supervision, while UPR operates within the model; both variants are evaluated on raw images.

We evaluate the resulting capabilities on USI-Bench, a held-out set of 6,525 QA pairs, and three external benchmarks. GeoQA improves depth and ordering, HUE improves localization and structural perception, and UPR produces its largest incremental gain in scene description. Table~\ref{tab:main-results} provides task-level component ablations, and Appendix~\ref{app:data-component-ablations} examines how training-group coverage changes the five capability scores. Object and ecological understanding provide domain context alongside the explicit 2D and 3D reasoning tasks.

The contributions are:
\begin{itemize}
    \item A spatial data engine that connects enhanced underwater imagery, instance annotations, and geometric cues to construct 4.27M QA pairs across five capability groups.
    \item USI-Brain, an underwater vision-language model trained on this corpus, and UPR, a prior-guided feature rectifier evaluated through matched-data training and component ablations.
    \item USI-Bench, a 6,525-QA benchmark that measures spatial reasoning and underwater semantics at the task level, complemented by evaluations on NautData, UWBench, and MarineEval.
\end{itemize}

\begin{table*}[!tp]
\centering
\small
\setlength{\tabcolsep}{4.0pt}
\caption{Comparison of training resources for underwater spatial intelligence. OG\&U denotes object grounding and understanding; the remaining groups cover 2D relations and layout, 3D depth and stratification, degradation-aware perception, and scene and ecological reasoning. Reported sizes use the counting units of the respective resources; the USI entry counts generated QA pairs.}
\label{tab:train-resource-comparison}
\begin{adjustbox}{max width=\linewidth}
\begin{tabular*}{\linewidth}{@{\extracolsep{\fill}}lcccccccc@{}}
\toprule
Training resource & Ref. & OG\&U & \shortstack{2D Rel.\\\&Layout} & \shortstack{3D Depth\\\&Strat.} & \shortstack{Deg.\\\&Perc.} & \shortstack{Scene\\\&Eco.} & \shortstack{Reported\\size} & Open src. \\
\midrule
MarineInst~\citep{zheng2024marineinst} & ECCV 2024 & $\checkmark$ & $\times$ & $\times$ & $\times$ & $\times$ & 2.20M & Part. Avail. \\
CoralSCOP~\citep{zheng2024coralmask} & CVPR 2024 & $\checkmark$ & $\times$ & $\times$ & $\times$ & $\sim$ & 46.60K & Avail. \\
SynTIDE~\citep{lin2025syntide} & CVPR 2025 & $\sim$ & $\times$ & $\checkmark$ & $\times$ & $\times$ & 54.51K & Avail. \\
NautData~\citep{xu2025nautilus} & NeurIPS 2025 & $\checkmark$ & $\sim$ & $\times$ & $\sim$ & $\sim$ & 1.45M & Avail. \\
\midrule
USI (ours) & --- & $\checkmark$ & $\checkmark$ & $\checkmark$ & $\checkmark$ & $\checkmark$ & 4.27M & Planned \\
\bottomrule
\end{tabular*}
\end{adjustbox}\par
\vspace{2pt}
{\raggedright \footnotesize \textit{Notation.} $\checkmark$: explicit support; $\sim$: partial or indirect support; $\times$: little or no direct support.\par}
\end{table*}

\begin{table*}[!tp]
\centering
\small
\setlength{\tabcolsep}{2.0pt}
\caption{Coverage of public evaluation resources across five underwater spatial-intelligence capability groups.}
\label{tab:eval-comparison}
\begin{adjustbox}{max width=\linewidth}
\begin{tabular}{@{}llccccccccccc@{}}
\toprule
\multirow{2}{*}{Evaluation resource} & \multirow{2}{*}{Ref.} & \multicolumn{3}{c}{OG\&U} & \multicolumn{2}{c}{2D Rel.\&Layout} & \multicolumn{2}{c}{3D Depth\&Strat.} & \multicolumn{2}{c}{Deg.\&Perc.} & \multicolumn{2}{c}{Scene\&Eco.} \\
\cmidrule(lr){3-5} \cmidrule(lr){6-7} \cmidrule(lr){8-9} \cmidrule(lr){10-11} \cmidrule(lr){12-13}
 &  & Obj. & Attr. & Grd./Cnt./Inst. & Rel. & Layout & Depth & Occ./View & Env. & Struct. & Scene & Eco. \\
\midrule
CoralVQA~\citep{han2025coralvqa} & arXiv 2025 & $\checkmark$ & $\checkmark$ & $\checkmark$ & $\sim$ & $\sim$ & $\times$ & $\times$ & $\times$ & $\times$ & $\sim$ & $\checkmark$ \\
UWBench~\citep{zhang2025uwbench} & arXiv 2025 & $\checkmark$ & $\checkmark$ & $\checkmark$ & $\sim$ & $\sim$ & $\times$ & $\times$ & $\sim$ & $\sim$ & $\sim$ & $\sim$ \\
MarineEval~\citep{wong2026marineeval} & WACV 2026 & $\checkmark$ & $\checkmark$ & $\checkmark$ & $\sim$ & $\sim$ & $\sim$ & $\sim$ & $\times$ & $\times$ & $\sim$ & $\checkmark$ \\
CoralSCOP (test split)~\citep{zheng2024coralmask} & CVPR 2024 & $\checkmark$ & $\checkmark$ & $\sim$ & $\times$ & $\times$ & $\times$ & $\times$ & $\times$ & $\times$ & $\times$ & $\sim$ \\
NautData (test split)~\citep{xu2025nautilus} & NeurIPS 2025 & $\checkmark$ & $\checkmark$ & $\checkmark$ & $\sim$ & $\sim$ & $\times$ & $\times$ & $\sim$ & $\sim$ & $\sim$ & $\sim$ \\
\midrule
USI-Bench (ours) & --- & $\checkmark$ & $\checkmark$ & $\checkmark$ & $\checkmark$ & $\checkmark$ & $\checkmark$ & $\checkmark$ & $\checkmark$ & $\checkmark$ & $\checkmark$ & $\checkmark$ \\
\bottomrule
\end{tabular}
\end{adjustbox}\par
\vspace{2pt}
{\raggedright \footnotesize \textit{Notation.} $\checkmark$: explicit coverage; $\sim$: partial coverage; $\times$: little or no targeted coverage.\par}
\end{table*}

\section{Related Work}

\textbf{Underwater Enhancement and 3D Supervision.}
Underwater enhancement methods address the appearance changes introduced by absorption, scattering, and illumination. Sea-Thru~\citep{akkaynak2019seathru} models underwater image formation, and UIEB~\citep{li2020uieb} provides an enhancement benchmark. Surveys describe the implications of these degradations for underwater perception and computational imaging \citep{huy2023underwaterperception,gu2023underwatercomputationalimaging}. Underwater depth estimation supplies geometric cues for object placement and scene structure \citep{cai2025underwaterdepth}. We bring these two sources of information into QA construction: enhancement prepares the visual evidence, and depth or reconstruction links it to object-level spatial relations.

\textbf{Spatial Reasoning Benchmarks and Data Engines.}
Spatial benchmarks measure object relationships, layouts, and geometric reasoning. Representative datasets include VSR~\citep{liu2023vsr}, SpaCE-10~\citep{gong2025space10}, SpatialBench~\citep{xu2025spatialbench}, OmniSpatial~\citep{jia2025omnispatial}, MM-Spatial~\citep{daxberger2025mmspatial}, InternSpatial~\citep{huang2025internspatial}, MMSI-Bench~\citep{yang2025mmsi}, and TangramPuzzle~\citep{liu2026tangram}. Related analyses study spatial generalization and embodied reasoning \citep{zhang2026generalization,yang2025embodiedbench}. SpatialVLM~\citep{chen2024spatialvlm}, Visual Spatial Tuning~\citep{yang2025vst}, SpatialLadder~\citep{liu2025spatialladder}, OpenSpatial~\citep{liu2026openspatial}, and Scaling Spatial Intelligence with Multimodal Foundation Models~\citep{zhou2025sensenovasi} investigate training with structured spatial data. EarthSpatialBench~\citep{xu2026earthspatialbench} extends spatial evaluation to a specific visual domain. Our engine uses object-linked spatial supervision in underwater scenes, combining the available geometric evidence with enhancement during question generation.

\textbf{Underwater Multimodal Models and Resources.}
NAUTILUS~\citep{xu2025nautilus} combines underwater data construction with multimodal model training. OceanGPT~\citep{bi2023oceangpt}, OceanGym~\citep{xue2025oceangym}, and USIM/U0~\citep{gu2025usimu0} address ocean-domain modeling, evaluation, or embodied interaction. General-purpose backbones such as Qwen3-VL~\citep{bai2025qwen3vl} provide a starting point for domain adaptation. USI-Brain builds on this resource-oriented approach by organizing enhancement, instance annotations, depth, and QA around a common set of underwater spatial capabilities.

\section{Methodology}
\label{sec:method}

Our method couples supervision construction with visual-feature adaptation. The data engine in Figure~\ref{fig:pipeline-overview} collects sources, enhances images, generates spatial QA, organizes capabilities, and filters annotations through expert review. Its 2D and geometry-assisted branches produce a shared training corpus. UPR then conditions visual-feature correction on underwater priors within the adapted model.

\subsection{Spatial Data Engine}

The engine accepts labeled underwater images, image-depth pairs, and frames extracted from underwater videos. Both QA branches use enhanced images. The 2D branch generates questions from image content and available annotations; the geometry branch combines images with object-aligned depth or reconstruction cues. Candidate questions are assigned to the five capability groups and passed to expert review for selection and refinement.

\textbf{Hybrid Underwater Enhancement.}
HUE prepares images for QA generation using two enhancement paths. The design is motivated by underwater image formation and restoration \citep{akkaynak2019seathru,li2020uieb,gu2023underwatercomputationalimaging}. The color-correction path predicts per-channel gains and biases from learned degradation maps and applies spatial refinement to adjust color and illumination. The local path uses a degradation-guided convolutional encoder--decoder with difference-aware enhancement to recover edges and textures.

Given an image \(I\in\mathbb{R}^{H\times W\times 3}\), an estimator predicts maps for color shift, scattering, and blur:
\begin{equation}
M_{\mathrm{deg}} =
\operatorname{Concat}(M_{\mathrm{color}},M_{\mathrm{scatter}},M_{\mathrm{blur}}).
\label{eq:degradation-maps}
\end{equation}
The learned maps condition both paths, producing a color-corrected image \(I_{\mathrm{cc}}\) and a detail-restored image \(I_{\mathrm{detail}}\). A fusion module takes \(I\) and \(M_{\mathrm{deg}}\) and predicts a pixel-wise weight map \(\alpha\in[0,1]^{H\times W}\):
\begin{equation}
I_{\mathrm{hue}} =
\alpha\odot I_{\mathrm{cc}}+(1-\alpha)\odot I_{\mathrm{detail}}.
\label{eq:hue-fusion}
\end{equation}
Here \(\odot\) denotes element-wise multiplication, with the spatial weights broadcast over image channels. The pixel-wise weights adapt the balance between color correction and detail restoration across the image.


\textbf{2D Spatial QA Generation.}
The 2D branch uses enhanced images and available annotations to generate questions about object identity and attributes, image-plane locations, and relationships between objects. It also provides questions about visibility and scene content. Each question is paired with its reference answer in the shared QA format used by both branches.

\textbf{3D Geometry Lifting and Reconstruction-Assisted Supervision.}
The geometry branch combines an enhanced RGB image with its corresponding depth map. Following reconstruction-assisted spatial supervision~\citep{liu2026openspatial}, we reconstruct Atlantis scenes~\citep{yang2025atlantis} and fit object-aligned 3D boxes. These boxes connect image regions to geometric measurements used for question generation.

For an instance mask or 2D bounding box, we extract the associated depth samples. A pixel \(u_i=(x_i,y_i,1)^\top\) with depth \(z_i\) is back-projected into camera coordinates:
\begin{equation}
p_i=z_iK^{-1}u_i,
\label{eq:backprojection}
\end{equation}
where \(K\in\mathbb{R}^{3\times 3}\) is the camera intrinsic matrix. We fit a 3D box to the resulting object point set \(\mathcal{P}_o=\{p_i\}_{i=1}^{n_o}\):
\begin{equation}
\mathcal{B}_o=\operatorname{FitBox}(\mathcal{P}_o).
\label{eq:box-fitting}
\end{equation}
The fitted boxes and reconstructed scenes provide cues for relative depth, foreground-background organization, occlusion, object-to-object distance, and viewpoint. Renderings with aligned box cues are supplied to the QA-generation model together with enhanced images. Depth maps, 3D boxes, and reconstructed scenes supply offline supervision; evaluation requires images and questions.


\textbf{Capability Organization and Expert Review.}
We organize candidate QA pairs into object grounding and understanding, 2D relations and layout, 3D depth and stratification, degradation-aware perception, and scene and ecological reasoning. Review checks visual evidence for the answer, consistency between questions, answers, and localization annotations, and redundancy across tasks. It selects and refines candidate annotations before the training corpus is assembled. The independent USI-Bench audit is described in Section~\ref{sec:benchmark}.


\begin{strip}
  \centering
  \includegraphics[width=\textwidth,height=0.38\textheight,keepaspectratio]{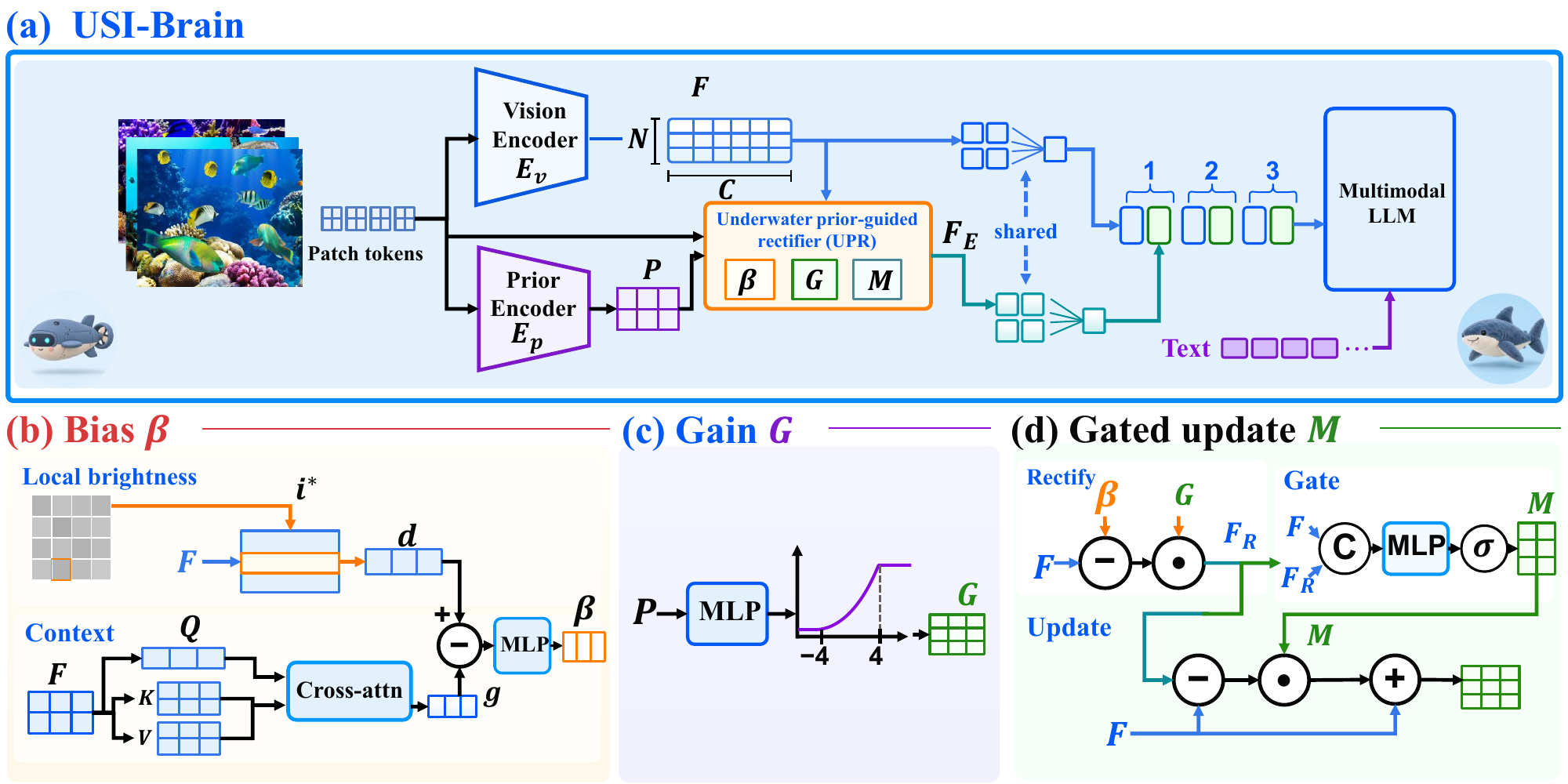}
  \captionof{figure}{Overview of UPR, which estimates scattering-bias and absorption-recovery signals and rectifies visual tokens through gated residual fusion.}
  \label{fig:upr-pipeline}
  \par\smallskip\raggedright
  \noindent\begin{minipage}[t]{\columnwidth}
    \subsection{Feature-space Rectification with UPR}
    \label{sec:upr}
  \end{minipage}\par
  \vspace{-\stripsep}
\end{strip}

\noindent USI-Brain + UPR uses UPR to adjust visual features extracted from degraded inputs; the USI-Brain variant omits this module. The module is placed after the vision encoder and before multimodal projection, while retaining the backbone ViT, merger, and language-model architectures. Figure~\ref{fig:upr-pipeline} shows its prior extraction, feature correction, and residual fusion steps.

Let \(F=\mathcal{E}_v(I)\in\mathbb{R}^{N\times C}\) denote visual tokens from the Qwen3-VL vision encoder, and let \(P\in\mathbb{R}^{N\times C}\) denote the aligned prior features derived from \(\mathcal{E}_p(I)\). The prior encoder \(\mathcal{E}_p\) is DINOv2 ViT-L/14. Motivated by the additive and multiplicative effects of underwater degradation \citep{gu2023underwatercomputationalimaging,huy2023underwaterperception}, UPR predicts a scattering-related bias \(\beta\) and an absorption-recovery gain \(G\):
\begin{equation}
\begin{aligned}
\beta &= \phi_s(d-g),\\
G &= \exp\!\bigl(\operatorname{clip}(\phi_a(P),-4,4)\bigr).
\end{aligned}
\label{eq:upr-corrections}
\end{equation}
Here \(d\) is the token selected from the darkest image region, \(g\) is a learned global context feature, and \(\phi_s\) and \(\phi_a\) are learned predictors. These terms parameterize corrections in feature space. The intermediate representation is
\begin{equation}
\tilde{F}=G\odot(F-\beta).
\label{eq:upr-intermediate}
\end{equation}

A gate controls how much of this correction is applied:
\begin{equation}
\begin{aligned}
M &= \sigma\!\bigl(\phi_g([F;\tilde{F}])\bigr),\\
\hat{F} &= F+M\odot(\tilde{F}-F).
\end{aligned}
\label{eq:upr-gate}
\end{equation}
The operator \([F;\tilde{F}]\) denotes channel-wise concatenation, \(\sigma\) is the sigmoid function, and \(\phi_g\) predicts the gate. The original and corrected features pass through the same merger. Let \(R,\hat{R}\in\mathbb{R}^{L\times D}\) be the two post-merger sequences. Their tokens are interleaved along the sequence dimension:
\begin{equation}
Z=\operatorname{Interleave}(R,\hat{R})\in\mathbb{R}^{2L\times D}.
\label{eq:upr-token-interleave}
\end{equation}
UPR preserves the channel width \(D\) and doubles the post-merger visual sequence. In the fixed \(512\times512\) configuration, 256 baseline visual tokens become 512. The gate weights the feature update element-wise, while interleaving presents the raw and corrected tokens as separate positions to the language model. UPR's inference computation includes the prior encoder and this longer visual sequence.


\subsection{USI-Brain Adaptation and Fine-tuning}

USI-Brain is initialized from Qwen3-VL-8B-Instruct~\citep{bai2025qwen3vl}. Training combines the 2D and geometry-assisted QA pairs in a single corpus, following the use of structured spatial supervision in prior work \citep{chen2024spatialvlm,yang2025vst,liu2026openspatial}. We use scaled dot-product attention (SDPA), unfreeze the ViT, language model, and merger, and train on 4.27M QA pairs for two epochs. Optimization settings are given in Section~\ref{sec:experimental-setup}.

We train USI-Brain without UPR and USI-Brain + UPR as separate checkpoints using the same corpus and optimization settings. The learning objective is cross-entropy over the QA answer tokens:
\begin{equation}
\mathcal{L}_{\mathrm{QA}}=-\sum_{i=1}^{|a|}\log p_\theta(a_i\mid I,q,a_{<i}),
\label{eq:qa-objective}
\end{equation}
where \(q\) and \(a\) denote the question and reference answer. In USI-Brain + UPR, the rectifier is optimized jointly with the backbone through this objective. Both variants optimize the QA objective without auxiliary distillation. Closed models generate and check QA offline. Component removals and restricted training-group configurations are defined in Appendix~\ref{app:data-component-ablations}.

\section{Dataset and Benchmark}
\label{sec:data}

The data engine constructs the training corpus, and USI-Bench is a held-out evaluation set. Splits are defined by original image or video: frames and processed derivatives remain in the same source split, together with their associated QA. Test-source images from USI-Bench, NautData, UWBench, and MarineEval are excluded from training-QA construction. Source IDs and QA links are checked, with SHA-256 and perceptual-hash screening for duplicate images. NautData is evaluated on its official within-dataset split; its 3,920 test images are excluded from training.

\subsection{Underwater Spatial Intelligence Dataset}

We use public underwater resources as visual sources and regenerate spatial QA supervision through the data engine. The source pool contains approximately 540K visual samples from still images and video frames, drawn from the ten datasets aggregated in NautData, together with CoralVQA, WebUOT-1M~\citep{zhang2024webuot}, and our reconstructions of approximately 400 Atlantis scenes. These sources cover different marine environments, objects, habitats, visibility conditions, and annotation formats. Table~\ref{tab:train-resource-comparison} compares their capability coverage.


The 2D branch generates QA pairs from enhanced images and annotations. The geometry branch adds questions that use depth and reconstructed scene structure, including depth ordering, stratification, occlusion, distance, and viewpoint. Capability-based organization and expert filtering produce a final corpus of 4.27M QA pairs.

\subsection{USI-Bench Composition and Evaluation}
\label{sec:benchmark}

\begin{figure*}[!tp]
  \centering
  \includegraphics[width=\textwidth,height=0.34\textheight,keepaspectratio]{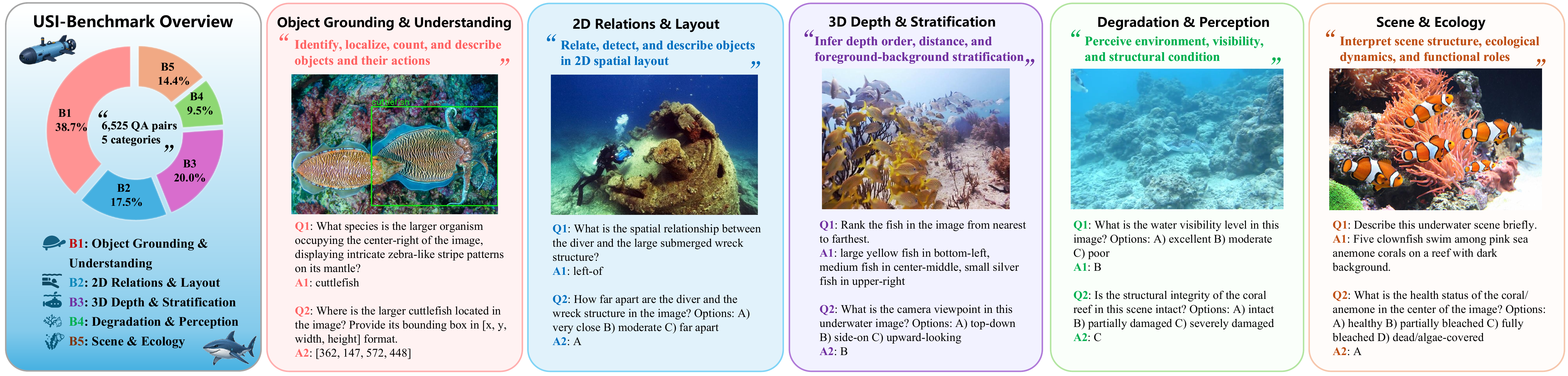}
  \caption{Composition and representative QA examples of USI-Bench across five capability groups.}
  \label{fig:benchmark-overview}
\end{figure*}

USI-Bench contains 6,525 held-out QA pairs drawn from public underwater resources and newly curated annotations. Training and benchmark QA share a generation--filtering--review workflow but use different prompts and templates. Fifteen independent annotators audited all 6,525 benchmark QA pairs. In the pre-revision audit, approximately 83\% were judged correct, 10\% uncertain, and 7\% incorrect. All uncertain and incorrect entries were subsequently revised. Figure~\ref{fig:benchmark-overview} shows the composition and examples. The benchmark covers five groups:

\begin{itemize}
    \item \textbf{Object Grounding and Understanding:} object identity, attributes, counting, actions, grounding, instance localization, and region descriptions.
    \item \textbf{2D Relations and Layout:} relative position, pairwise relationships, local arrangement, and coarse distance.
    \item \textbf{3D Depth and Stratification:} relative depth, ordering, foreground-background separation, occlusion, and viewpoint.
    \item \textbf{Degradation-aware Perception:} visibility and structural cues affected by color cast, blur, low illumination, and turbidity.
    \item \textbf{Scene and Ecological Reasoning:} scene descriptions, habitat cues, ecological relationships, health status, and the functions of underwater structures.
\end{itemize}

Tasks use class labels, multiple-choice answers, ranked outputs, or bounding boxes where direct scoring is possible \citep{liu2023vsr,gong2025space10,liu2026openspatial}. Table~\ref{tab:main-results} reports accuracy for classification, Acc@IoU0.5 for grounding, Recall@IoU0.5 for detection, Kendall's \(\tau\) for ordering, and macro-F1 for foreground-background separation and structural perception. Grounding outputs use integer \(xywh\) coordinates normalized to 0--1000. Metrics are displayed on a 100-point scale, including \(100\tau\) for ordering.

For foreground-background separation, option letters are extracted, deduplicated, and sorted; each complete answer combination is one class. Macro-F1 is the equal-weight mean of classwise F1 over the union of reference and predicted classes, multiplied by 100. The fixed-answer baseline that always predicts A scores 3.43 under this protocol.

Scene and region captioning use three independent judges: Gemini 3.1 Pro, GPT-5.5, and Claude Opus 4.7. Each receives the same question, reference answer, and model response, and assigns a score from 0 to 5 based on semantic agreement, with penalties for omissions and unsupported claims \citep{zheng2023llmjudge}. Scene scoring considers the overall scene, key entities, actions, and spatial relationships; caption scoring considers the described region's entities, attributes, counts, actions, and relationships. We average over the three judges and all items within each task, then multiply by 20.


\section{Experiments}

\subsection{Experimental Setup}
\label{sec:experimental-setup}

\textbf{Comparison Models.}
USI-Bench comparisons include proprietary models, open-source general VLMs, spatial-understanding models, and underwater-specialized models. The general-purpose models include Qwen3-VL~\citep{bai2025qwen3vl}, GPT-5.4, Gemini~\citep{google2025gemini25,google2026gemini31pro}, and Claude. The spatial models are SenseNova-SI~\citep{zhou2025sensenovasi} and VST~\citep{yang2025vst}; the underwater models are OceanGPT~\citep{bi2023oceangpt} and the Qwen2.5-VL- and LLaVA-1.5-based NAUTILUS variants~\citep{xu2025nautilus}. Additional Qwen2.5-VL~\citep{bai2025qwen25vl}, InternVL~\citep{chen2024internvl}, and GLM-V~\citep{hong2025glm4v} comparisons are included in the external results.


\textbf{Evaluation.}
For USI-Bench, models are evaluated through a common prompting, decoding, and answer-normalization interface. Predictions are parsed into task-specific outputs before scoring. The external benchmarks retain the comparison protocols described in Appendix~\ref{app:external-results}, including published baseline scores on UWBench and a separate three-judge evaluation on MarineEval.

\textbf{Training.}
We fine-tune Qwen3-VL-8B-Instruct on the 4.27M-QA corpus for two epochs. The language and vision components use learning rates of \(5\times10^{-5}\) and \(5\times10^{-6}\), respectively, with a warmup ratio of 0.1 and cosine decay. Training uses mixed precision, gradient checkpointing, and FSDP full sharding. The micro-batch size is 2 and the global batch size is 128 on 64 NVIDIA B200 GPUs. The full training run takes approximately 25 hours. Frequency-domain gating in HUE is disabled in all reported experiments.

\textbf{Model Variants and Ablations.}
USI-Brain omits UPR; USI-Brain + UPR includes it. They are trained separately on the same 4.27M QA pairs with the same training settings. For USI-Bench evaluation, both use raw images with identical processor settings and dynamic resizing; HUE is not applied at inference. In Table~\ref{tab:main-results}, \emph{w/o HUE} removes enhancement-derived supervision while retaining GeoQA, and \emph{w/o GeoQA} removes the geometry-assisted QA branch while retaining HUE. Both removals keep UPR disabled and are compared with USI-Brain. Appendix~\ref{app:data-component-ablations} specifies their correspondence to the training-group configurations and reports the extended data ablations.


Single-GPU inference latency, decoding speed, and peak memory on A800 and RTX 3090 are reported in Appendix~\ref{app:inference-efficiency}.

\subsection{Main Results on USI-Bench}

\begin{table*}[!t]
\centering
\footnotesize
\renewcommand{\arraystretch}{1.0}
\setlength{\tabcolsep}{2.6pt}
\caption{Fine-grained results on USI-Bench. Metrics are reported on a 100-point scale, with Kendall's $\tau$ multiplied by 100. Parentheses show score differences from Qwen3-VL-8B-Instruct. Boldface marks the best or joint-best result among the listed models. Within Ours, darker/lighter blue marks the best/second-best distinct scores; ties share a color. USI-Brain uses HUE-derived and geometry-assisted QA (GeoQA) supervision without UPR; USI-Brain + UPR is trained with UPR. Both removal ablations disable UPR and are compared with USI-Brain. All variants use raw evaluation images. Positive/negative differences are red/green.}
\label{tab:main-results}
\begin{adjustbox}{max width=\linewidth}
\begin{tabular*}{\linewidth}{@{\extracolsep{\fill}}lccccccc@{}}
\toprule
Model & \multicolumn{5}{c}{Object Grounding \& Understanding} & \multicolumn{2}{c}{2D Relations \& Layout} \\
\cmidrule(lr){2-6} \cmidrule(lr){7-8}
 & Identity & Grounding & Counting & Detection & Captioning & Relation & Distance \\
\cmidrule(lr){2-2} \cmidrule(lr){3-3} \cmidrule(lr){4-4} \cmidrule(lr){5-5} \cmidrule(lr){6-6} \cmidrule(lr){7-7} \cmidrule(lr){8-8}
 & Acc & Acc@IoU0.5 & Acc & Recall@IoU0.5 & Judge & Acc & Acc \\
\midrule
\multicolumn{8}{@{}l}{\textit{Proprietary Models}} \\
GPT-5.4 & 50.5 & 49.6 & 46.3 & 8.9 & 53.2 & 60.5 & 75.2 \\
Gemini & 50.1 & 8.7 & 45.0 & 0.1 & 45.4 & \textbf{73.0} & \textbf{77.0} \\
Claude-Opus-4.6 & 60.3 & 28.9 & 42.9 & 7.0 & \textbf{57.7} & 40.0 & 57.5 \\
\midrule
\multicolumn{8}{@{}l}{\textit{Open-source General VLMs}} \\
Qwen3-VL-30B-A3B-Instruct & 67.6 & 86.4 & 46.5 & 49.4 & 31.3 & 68.5 & 72.6 \\
Qwen3-VL-8B-Instruct & 64.7 & 85.4 & 42.9 & 47.9 & 38.9 & 65.8 & 69.9 \\
\midrule
\multicolumn{8}{@{}l}{\textit{Spatial-understanding Models}} \\
SenseNova-SI-1.1-Qwen3-VL-8B & 59.5 & 82.8 & 39.6 & 32.8 & 34.9 & 60.6 & 60.2 \\
VST-7B-SFT & 54.4 & 76.2 & \textbf{47.5} & 35.1 & 5.3 & 44.5 & 56.3 \\
\midrule
\multicolumn{8}{@{}l}{\textit{Underwater-specialized Models}} \\
OceanGPT & 57.9 & 1.4 & 46.5 & 0.6 & 38.8 & 42.8 & 45.4 \\
NAUTILUS-Qwen2.5-VL & 63.2 & 0.2 & 35.5 & 0.3 & 43.3 & 57.8 & 60.8 \\
NAUTILUS-LLaVA-1.5 & 52.6 & 0.6 & 38.7 & 0.1 & 41.5 & 53.5 & 55.2 \\
\midrule
\multicolumn{8}{@{}l}{\textit{Ours}} \\
USI-Brain + UPR & \bestcell \textbf{73.7} \gain{+9.0} & \bestcell \textbf{92.1} \gain{+6.7} & \bestcell 44.2 \gain{+1.3} & \bestcell \textbf{56.3} \gain{+8.4} & \bestcell 56.2 \gain{+17.3} & \bestcell 72.8 \gain{+7.0} & \bestcell 69.6 \drop{-0.3} \\
USI-Brain & 73.1 \gain{+8.4} & \secondcell 90.8 \gain{+5.4} & 42.4 \drop{-0.5} & \secondcell 55.5 \gain{+7.6} & \secondcell 55.9 \gain{+17.0} & \secondcell 71.4 \gain{+5.6} & \secondcell 69.0 \drop{-0.9} \\
USI-Brain w/o HUE & 70.6 \gain{+5.9} & 88.2 \gain{+2.8} & 41.0 \drop{-1.9} & 51.9 \gain{+4.0} & 52.5 \gain{+13.6} & 70.5 \gain{+4.7} & 68.7 \drop{-1.2} \\
USI-Brain w/o GeoQA & \secondcell 73.2 \gain{+8.5} & 89.6 \gain{+4.2} & \secondcell 43.8 \gain{+0.9} & 54.2 \gain{+6.3} & 54.9 \gain{+16.0} & 67.6 \gain{+1.8} & 66.1 \drop{-3.8} \\
\bottomrule
\end{tabular*}
\end{adjustbox}\par

\vspace{0.2em}

\begin{adjustbox}{max width=\linewidth}
\begin{tabular*}{\linewidth}{@{\extracolsep{\fill}}lcccccccc@{}}
\toprule
Model & \multicolumn{4}{c}{3D Depth \& Strat.} & \multicolumn{2}{c}{Degradation \& Perception} & \multicolumn{2}{c}{Scene \& Ecology} \\
\cmidrule(lr){2-5} \cmidrule(lr){6-7} \cmidrule(lr){8-9}
 & Depth & Viewpoint & Ordering & Fg./Bg. & Environment & Structure & Scene & Ecology \\
\cmidrule(lr){2-2} \cmidrule(lr){3-3} \cmidrule(lr){4-4} \cmidrule(lr){5-5} \cmidrule(lr){6-6} \cmidrule(lr){7-7} \cmidrule(lr){8-8} \cmidrule(lr){9-9}
 & Acc & Acc & $100\tau$ & macro-F1 & Acc & macro-F1 & Judge & Acc \\
\midrule
\multicolumn{9}{@{}l}{\textit{Proprietary Models}} \\
GPT-5.4 & 58.4 & 83.3 & 76.4 & 19.7 & 82.7 & 80.9 & 68.0 & \textbf{75.0} \\
Gemini & 61.7 & 77.2 & 80.2 & 19.2 & 76.5 & \textbf{83.7} & \textbf{68.2} & 64.7 \\
Claude-Opus-4.6 & 24.0 & 71.9 & 79.5 & 18.0 & 60.2 & 25.5 & 25.9 & 44.2 \\
\midrule
\multicolumn{9}{@{}l}{\textit{Open-source General VLMs}} \\
Qwen3-VL-30B-A3B-Instruct & 77.5 & 82.0 & 80.9 & 19.0 & \textbf{84.9} & 61.1 & 43.3 & 61.3 \\
Qwen3-VL-8B-Instruct & 76.0 & 79.1 & 72.2 & 18.5 & 81.2 & 46.5 & 42.8 & 58.4 \\
\midrule
\multicolumn{9}{@{}l}{\textit{Spatial-understanding Models}} \\
SenseNova-SI-1.1-Qwen3-VL-8B & 76.0 & 73.6 & 66.2 & 19.2 & 76.3 & 51.6 & 39.5 & 53.5 \\
VST-7B-SFT & 44.6 & 54.8 & 70.5 & 19.2 & 71.6 & 42.1 & 29.7 & 44.6 \\
\midrule
\multicolumn{9}{@{}l}{\textit{Underwater-specialized Models}} \\
OceanGPT & 65.3 & 71.0 & 52.8 & 18.7 & 62.7 & 19.7 & 20.3 & 56.4 \\
NAUTILUS-Qwen2.5-VL & 58.7 & 78.4 & 40.1 & 15.9 & 72.6 & 56.7 & 24.7 & 54.8 \\
NAUTILUS-LLaVA-1.5 & 60.5 & 42.9 & 20.9 & 10.0 & 71.6 & 59.2 & 61.9 & 47.8 \\
\midrule
\multicolumn{9}{@{}l}{\textit{Ours}} \\
USI-Brain + UPR & \bestcell \textbf{83.8} \gain{+7.8} & \bestcell \textbf{84.8} \gain{+5.7} & \bestcell \textbf{83.2} \gain{+11.0} & \bestcell \textbf{22.7} \gain{+4.2} & \bestcell 78.3 \drop{-2.9} & \bestcell 62.5 \gain{+16.0} & \bestcell 65.3 \gain{+22.5} & \bestcell 59.1 \gain{+0.7} \\
USI-Brain & \secondcell 82.7 \gain{+6.7} & \bestcell \textbf{84.8} \gain{+5.7} & \secondcell 82.3 \gain{+10.1} & \secondcell 22.6 \gain{+4.1} & \secondcell 77.8 \drop{-3.4} & \secondcell 62.2 \gain{+15.7} & \secondcell 47.3 \gain{+4.5} & \secondcell 58.8 \gain{+0.4} \\
USI-Brain w/o HUE & 80.8 \gain{+4.8} & \secondcell 84.1 \gain{+5.0} & 80.6 \gain{+8.4} & 21.4 \gain{+2.9} & 76.8 \drop{-4.4} & 58.9 \gain{+12.4} & 45.1 \gain{+2.3} & 58.3 \drop{-0.1} \\
USI-Brain w/o GeoQA & 77.9 \gain{+1.9} & 81.6 \gain{+2.5} & 76.2 \gain{+4.0} & 20.4 \gain{+1.9} & 77.7 \drop{-3.5} & 60.4 \gain{+13.9} & 46.7 \gain{+3.9} & 58.6 \gain{+0.2} \\
\bottomrule
\end{tabular*}
\end{adjustbox}\par
\end{table*}

Table~\ref{tab:main-results} reports 15 task metrics across five capability groups, including the two component ablations. USI-Brain improves over Qwen3-VL-8B-Instruct on 12 of 15 metrics. The largest gains are in captioning (+17.0 points), structure (+15.7), and ordering (+10.1 on the \(100\tau\) scale). Grounding rises from 85.4 to 90.8, and detection recall from 47.9 to 55.5. This UPR-free model captures the gains from fine-tuning on the constructed corpus.

USI-Brain + UPR improves 13 of 15 metrics over the backbone and achieves the best or joint-best result among the listed models on seven: identity, grounding, detection, depth, viewpoint, ordering, and foreground-background separation. It exceeds the non-USI open-source comparison models on 11 metrics. Its four depth and stratification scores are 83.8, 84.8, 83.2, and 22.7, respectively.

Averaging the signed score differences across all 15 tasks gives gains of 5.76 points for USI-Brain and 7.63 for USI-Brain + UPR. Each task receives equal weight on its reported 100-point scale, including negative differences. Appendix~\ref{app:data-component-ablations} also reports a capability-level summary that gives each of the five groups equal weight, yielding gains of 5.04 and 7.44 points, respectively.

\subsection{Component and Training-Data Analysis}
\label{sec:ablation-analysis}

\textbf{Geometry-assisted supervision.}
With UPR disabled, retaining GeoQA raises Depth, Viewpoint, Ordering, and Fg./Bg. by 4.8, 3.2, 6.1, and 2.2 points, respectively. The largest gains occur in depth and ordering, the tasks directly supported by object-level geometry. Relation and Distance also improve by 3.8 and 2.9 points. This comparison varies the inclusion of geometry-assisted QA while retaining HUE, revealing benefits across both 3D and image-plane relationships.

\textbf{Enhancement-derived supervision.}
With GeoQA retained and UPR disabled, HUE improves all 15 task scores. Grounding, Depth, Ordering, Structure, and Scene rise by 2.6, 1.9, 1.7, 3.3, and 2.2 points, respectively. Enhancement during supervision construction thus improves localization and structural judgments on raw evaluation images.

\textbf{Feature rectification.}
Training with UPR on the same QA corpus adds 1.3 points in Grounding, 1.1 in Depth, and 0.9 in Ordering, with Viewpoint unchanged. Its largest effect is on Scene, which rises from 47.3 to 65.3 (+18.0). Captioning and Ecology each increase by 0.3. The gain is concentrated in global scene description, with smaller changes in region description and ecological classification. Both models use HUE-derived supervision.

\textbf{Training-data coverage.}
Table~\ref{tab:data-component-ablations} in Appendix~\ref{app:data-component-ablations} examines restricted training groups. Adding B3 to B1 training raises the B3 gain over Base from +0.50 to +4.60; adding B4 to B1+B2+B3 raises the B4 gain from +0.20 to +3.00. In these two comparisons, the largest change is in the newly included capability group. The four configurations shared with Table~\ref{tab:main-results} connect these group-level patterns to the task-level component results.

\textbf{Task-level variation.}
Distance and Environment remain below the backbone at 69.6 versus 69.9 and 78.3 versus 81.2, respectively. Fg./Bg. reaches 22.7, compared with 18.5 for the backbone and 3.43 for the fixed-A reference. Foreground-background assignment, distance estimation, and environment recognition thus remain priorities for expanding supervision and evaluation. Detection is measured by Recall@IoU0.5; precision would complement this measure by characterizing false positives.

\subsection{External Benchmark Evaluation}

\begin{figure}[!tbp]
  \centering
  \includegraphics[width=\columnwidth,keepaspectratio]{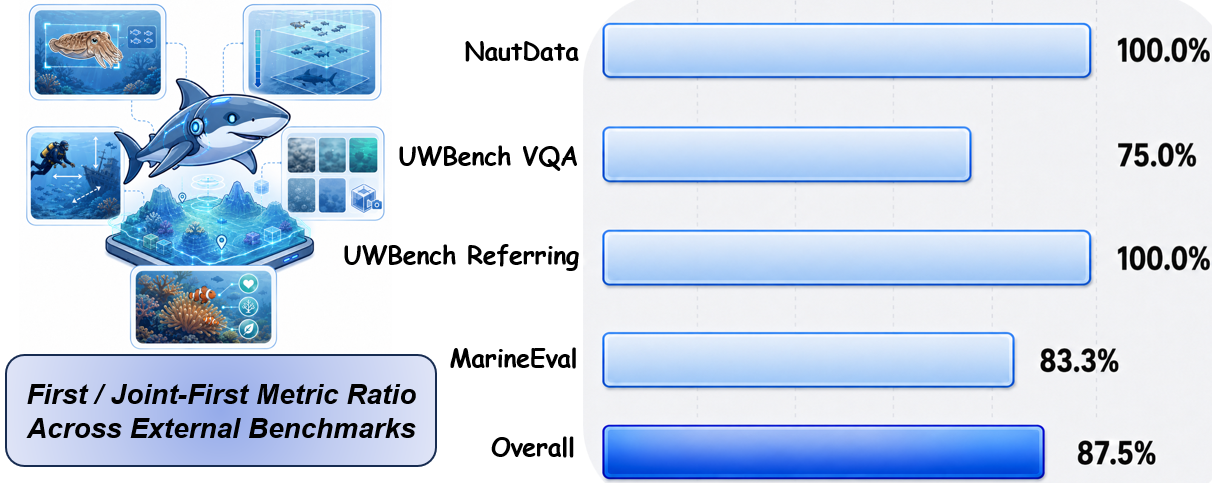}
  \caption{Percentage of reported metrics on which USI-Brain achieves the best or joint-best score within each comparison pool. UWBench VQA covers selected categories, and MarineEval uses the three-judge protocol described in Appendix~\ref{app:external-results}.}
  \label{fig:external-benchmark-scores}
\end{figure}

We evaluate USI-Brain on NautData~\citep{xu2025nautilus}, UWBench~\citep{zhang2025uwbench}, and MarineEval~\citep{wong2026marineeval}. Figure~\ref{fig:external-benchmark-scores} summarizes its best-or-joint-best counts: 7/7 on NautData, 6/8 on the selected UWBench VQA categories, 5/5 on UWBench object referring, and 10/12 on MarineEval. Across these comparison pools, USI-Brain leads 27 metrics and ties on one, giving 28/32 (87.5\%). Its mean signed difference from the strongest baseline on each metric is +9.99 points after rescaling to 100-point scores; Appendix~\ref{app:external-summary} defines the aggregation.

The gains span multiple output formats. On NautData, VQA accuracy rises from 0.381 for the strongest baseline to 0.478, while grounding mIoU rises from 53.8 to 56.4. On UWBench object referring, Acc@IoU$_{0.7}$ improves by 13.87 points. MarineEval shows gains of 15 points in spatial-relation assessment and 12 in numerical estimation. Appendix~\ref{app:external-results} provides the full category results, published-baseline comparisons, and the separate three-judge protocol used for MarineEval.

\section{Conclusion}

USI-Brain connects underwater appearance and geometry through a spatial data engine, model adaptation, and task-level evaluation. The engine converts images, annotations, and geometric cues into 4.27M QA pairs; USI-Bench measures the resulting capabilities with 6,525 held-out questions. Fine-tuning on this corpus yields a 5.76-point mean gain across 15 tasks, increasing to 7.63 with UPR.

The ablations identify distinct roles for the components: geometry-assisted QA strengthens depth and ordering, HUE-derived supervision improves localization and structural perception, and UPR contributes most strongly to scene description. Training-group ablations in Appendix~\ref{app:data-component-ablations} show how supervision coverage shapes the capability profile, while external evaluations extend the evidence to underwater QA and localization on three benchmarks. We will release the model weights and USI-Bench for further evaluation and development.

{
\small
\bibliography{aaai2026}
}

\clearpage
\onecolumn
\appendix
\small
\captionsetup{font=small}
\setlength{\parskip}{2pt}

\section{External Benchmark Results}
\label{app:external-results}

Tables~\ref{tab:external-summary}--\ref{tab:marineeval-results} give aggregate and category-level results on NautData, UWBench, and MarineEval. The following subsections specify each comparison pool, metric, and evaluation protocol. In Tables~\ref{tab:nautdata-results}--\ref{tab:marineeval-results}, darker/lighter blue marks the best/second-best distinct score in each column among the listed models; ties share a color.

\subsection{Summary and Aggregation}
\label{app:external-summary}

Table~\ref{tab:external-summary} summarizes 32 reported metrics. USI-Brain achieves a strictly higher score than all compared baselines on 27 metrics and ties for the highest score on one, giving a best-or-joint-best rate of 87.5\%.

For aggregation, scores reported on a unit scale are multiplied by 100; scores already on a 100-point scale are left unchanged. Let \(s_j\) denote the rescaled USI-Brain score on metric \(j\), and let \(b_{j,k}\) denote the corresponding score of baseline \(k\) in comparison pool \(\mathcal{C}_j\). We compute
\begin{equation}
\begin{aligned}
\Delta_j &= s_j-\max_{k\in\mathcal{C}_j}b_{j,k},\\
\overline{\Delta} &= \frac{1}{J}\sum_{j=1}^{J}\Delta_j.
\end{aligned}
\label{eq:external-aggregate}
\end{equation}
With \(J=32\), this gives \(\overline{\Delta}=9.99\) points. Every listed metric has equal weight, negative differences retain their signs, and CLAIR contributes its reported mean.

\par\medskip\noindent\begin{minipage}{\textwidth}
\centering
\small
\setlength{\tabcolsep}{10.0pt}
\captionof{table}{Summary of the reported external comparisons. \textit{Sole best} and \textit{joint best} compare USI-Brain with the strongest non-USI-Brain baseline for each metric. Mean score differences use Equation~\ref{eq:external-aggregate}, with all scores rescaled to a 100-point scale.}
\label{tab:external-summary}
\begin{adjustbox}{max width=\linewidth}
\begin{tabular*}{\linewidth}{@{\extracolsep{\fill}}lccc@{}}
\toprule
Benchmark & Sole best & Joint best & Mean score difference \\
\midrule
NautData~\citep{xu2025nautilus} & 7/7 & 0 & +2.77 points \\
UWBench VQA~\citep{zhang2025uwbench} & 6/8 & 0 & +24.41 points \\
UWBench Object Referring~\citep{zhang2025uwbench} & 5/5 & 0 & +7.80 points \\
MarineEval~\citep{wong2026marineeval} & 9/12 & 1/12 & +5.50 points \\
\midrule
\textbf{Overall} & \textbf{27/32} & \textbf{1/32} & \textbf{+9.99 points} \\
\bottomrule
\end{tabular*}
\end{adjustbox}\par
\end{minipage}\par\medskip

\Needspace{5\baselineskip}
\subsection{NautData}
\label{app:nautdata}

Table~\ref{tab:nautdata-results} reports results on the public NautData test split from NAUTILUS~\citep{xu2025nautilus}. The seven metrics cover image and region captioning, visual question answering, grounding, and detection. Captioning and VQA retain their unit-scale scores; Table~\ref{tab:external-summary} multiplies these by 100 when computing the aggregate.


\par\medskip\noindent\begin{minipage}{\textwidth}
\centering
\small
\setlength{\tabcolsep}{4.2pt}
\captionof{table}{Results on the public NautData test split from NAUTILUS~\citep{xu2025nautilus}. Image and region captioning use METEOR, VQA uses accuracy, grounding uses PR@0.5 and mIoU, and detection uses mAP and mAP@0.5. METEOR and VQA accuracy are shown on a unit scale; the remaining metrics are shown on a 100-point scale. Boldface marks the best score among the listed methods.}
\label{tab:nautdata-results}
\begin{adjustbox}{max width=\linewidth}
\begin{tabular*}{\linewidth}{@{\extracolsep{\fill}}lccccccc@{}}
\toprule
\multirow{2}{*}{Method} & Image Caption & Region Caption & \multirow{2}{*}{VQA} & \multicolumn{2}{c}{Grounding} & \multicolumn{2}{c}{Detection} \\
\cmidrule(lr){2-2} \cmidrule(lr){3-3} \cmidrule(lr){5-6} \cmidrule(lr){7-8}
 & METEOR & METEOR & Acc & PR@0.5 & mIoU & mAP & mAP@0.5 \\
\midrule
NAUTILUS (LLaVA-1.5)~\citep{xu2025nautilus} & 0.208 & 0.191 & 0.365 & 52.2 & 46.2 & 11.1 & 20.9 \\
NAUTILUS (Qwen2.5-VL)~\citep{xu2025nautilus} & \secondcell 0.223 & \secondcell 0.199 & \secondcell 0.381 & \secondcell 58.8 & \secondcell 53.8 & \secondcell 25.8 & \secondcell 45.3 \\
Qwen3-VL-8B~\citep{bai2025qwen3vl} & 0.089 & 0.097 & 0.008 & 45.2 & 43.9 & 14.8 & 24.0 \\
USI-Brain (ours) & \bestcell \textbf{0.245} & \bestcell \textbf{0.206} & \bestcell \textbf{0.478} & \bestcell \textbf{59.4} & \bestcell \textbf{56.4} & \bestcell \textbf{27.9} & \bestcell \textbf{46.8} \\
\bottomrule
\end{tabular*}
\end{adjustbox}\par
\end{minipage}\par\medskip

USI-Brain achieves the highest score among the listed methods on all seven metrics. VQA accuracy increases from 0.381 for the strongest baseline to 0.478, while grounding mIoU increases from 53.8 to 56.4. Image-caption METEOR increases from 0.223 to 0.245, and detection mAP increases from 25.8 to 27.9.

\Needspace{5\baselineskip}
\subsection{UWBench Visual Question Answering}
\label{app:uwbench-vqa}

Table~\ref{tab:uwbench-vqa-results} reports eight selected UWBench VQA categories related to object understanding, spatial reasoning, and underwater visibility. Baseline scores are taken from the UWBench paper~\citep{zhang2025uwbench}. USI-Brain is evaluated on the official test split using the same protocol.


\par\medskip\noindent\begin{minipage}{\textwidth}
\centering
\small
\setlength{\tabcolsep}{4.0pt}
\captionof{table}{Selected UWBench VQA results~\citep{zhang2025uwbench}. All scores use a 100-point scale. Baseline numbers are taken from the UWBench paper; USI-Brain is evaluated on the official test split under the same protocol. Boldface marks the best score among the listed methods.}
\label{tab:uwbench-vqa-results}
\begin{adjustbox}{max width=\linewidth}
\begin{tabular*}{\linewidth}{@{\extracolsep{\fill}}lcccc@{}}
\toprule
Method & \shortstack{Qty.\\\& Exist.} & \shortstack{Reas.\\\& Comp.} & \shortstack{Water Qual.\\\& Vis.} & \shortstack{Obj. Attr.\\\& Feat.} \\
\midrule
Gemini-2.5-Flash~\citep{google2025gemini25} & 1.47 & 26.14 & \secondcell 40.77 & 24.37 \\
Qwen2.5-VL-7B~\citep{bai2025qwen25vl} & \secondcell 49.91 & 23.24 & 35.54 & 24.37 \\
GLM-4.1V-9B~\citep{hong2025glm4v} & 2.33 & 26.97 & 39.94 & \secondcell 28.12 \\
GLM-4.5V-106B~\citep{hong2025glm4v} & 8.73 & \secondcell 29.05 & 40.22 & 27.50 \\
USI-Brain (ours) & \bestcell \textbf{92.72} & \bestcell \textbf{85.11} & \bestcell \textbf{82.50} & \bestcell \textbf{78.09} \\
\bottomrule
\end{tabular*}
\end{adjustbox}\par

\vspace{0.4em}

\begin{adjustbox}{max width=\linewidth}
\begin{tabular*}{\linewidth}{@{\extracolsep{\fill}}lcccc@{}}
\toprule
Method & \shortstack{Substrate\\\& Materials} & \shortstack{Pos.\\\& Spatial Rel.} & \shortstack{Color, Texture\\\& Pattern} & \shortstack{Shape, Size\\\& Form} \\
\midrule
Gemini-2.5-Flash~\citep{google2025gemini25} & 47.58 & \secondcell 61.55 & 45.26 & 71.42 \\
Qwen2.5-VL-7B~\citep{bai2025qwen25vl} & 42.24 & 55.66 & 38.95 & 64.42 \\
GLM-4.1V-9B~\citep{hong2025glm4v} & 47.17 & 61.27 & 45.02 & \secondcell 71.98 \\
GLM-4.5V-106B~\citep{hong2025glm4v} & \secondcell 49.85 & \bestcell \textbf{62.48} & \secondcell 48.78 & \bestcell \textbf{72.52} \\
USI-Brain (ours) & \bestcell \textbf{72.45} & 56.73 & \bestcell \textbf{56.33} & 52.79 \\
\bottomrule
\end{tabular*}
\end{adjustbox}\par
\end{minipage}\par\medskip

USI-Brain achieves the highest scores in six of the eight reported categories: quantity and existence, reasoning and comparison, water quality and visibility, object attributes and features, substrate and materials, and color, texture, and pattern. GLM-4.5V-106B scores higher on position and spatial relations (62.48 versus 56.73) and shape, size, and form (72.52 versus 52.79).

\Needspace{5\baselineskip}
\subsection{UWBench Object Referring}
\label{app:uwbench-referring}

Table~\ref{tab:uwbench-referring-results} reports four localization metrics from the official UWBench object-referring protocol and a complementary CLAIR score for open-ended output quality. The reported CLAIR means are used in the aggregate comparison.


\par\medskip\noindent\begin{minipage}{\textwidth}
\centering
\small
\setlength{\tabcolsep}{4.0pt}
\captionof{table}{Object-referring results on UWBench~\citep{zhang2025uwbench}. The first four metrics follow the official localization protocol. CLAIR is a complementary judge-based metric, reported as mean $\pm$ standard deviation. Scores use a 100-point scale. Baseline results are taken from the UWBench paper. Boldface marks the best reported mean or score in each column.}
\label{tab:uwbench-referring-results}
\begin{adjustbox}{max width=\linewidth}
\begin{tabular*}{\linewidth}{@{\extracolsep{\fill}}lccccc@{}}
\toprule
Method & Acc@IoU$_{0.5}$ & Acc@IoU$_{0.7}$ & mIoU & Cum\_IoU & CLAIR \\
\midrule
Qwen2.5-VL-7B~\citep{bai2025qwen25vl} & 37.15 & 11.76 & 32.77 & 33.69 & 87.28$\pm$4.86 \\
Qwen2.5-VL-72B~\citep{bai2025qwen25vl} & 90.90 & 74.71 & 77.90 & \secondcell 81.50 & 88.76$\pm$4.64 \\
InternVL-3.5-24B~\citep{chen2024internvl} & 84.76 & 64.14 & 71.99 & 76.98 & \secondcell 88.99$\pm$4.42 \\
Qwen3-VL-30B-Instruct~\citep{bai2025qwen3vl} & \secondcell 94.40 & \secondcell 80.38 & \secondcell 80.18 & 51.67 & 88.58$\pm$5.21 \\
GLM-4.5V-106B~\citep{hong2025glm4v} & 89.21 & 75.38 & 79.26 & 80.51 & 88.63$\pm$6.43 \\
USI-Brain (ours) & \bestcell \textbf{96.80} & \bestcell \textbf{94.25} & \bestcell \textbf{91.05} & \bestcell \textbf{92.07} & \bestcell \textbf{90.27$\pm$4.89} \\
\bottomrule
\end{tabular*}
\end{adjustbox}\par
\end{minipage}\par\medskip

USI-Brain achieves the highest values on all four localization metrics and the highest mean CLAIR score among the listed methods. The largest localization gain is at the stricter IoU threshold: Acc@IoU$_{0.7}$ increases from 80.38 for the strongest baseline to 94.25, a difference of 13.87 points. Its mIoU is 91.05, compared with 80.18 for the strongest baseline.

\Needspace{5\baselineskip}
\subsection{MarineEval}
\label{app:marineeval}

Table~\ref{tab:marineeval-results} reports 12 categories from MarineEval~\citep{wong2026marineeval}. All listed models are scored under our three-judge protocol using Gemini-3-Pro-Preview, GPT-5.5, and Claude Opus 4.7. The table reports this shared evaluation; the original MarineEval leaderboard uses a different protocol.


\par\medskip\noindent\begin{minipage}{\textwidth}
\centering
\small
\setlength{\tabcolsep}{3.8pt}
\captionof{table}{Selected MarineEval results under the reported three-judge protocol~\citep{wong2026marineeval}. Scores use a 100-point scale. The judges are Gemini-3-Pro-Preview, GPT-5.5, and Claude Opus 4.7. Boldface marks the best or joint-best score among the listed methods.}
\label{tab:marineeval-results}
\begin{adjustbox}{max width=\linewidth}
\begin{tabular*}{\linewidth}{@{\extracolsep{\fill}}lcccccc@{}}
\toprule
Method & \shortstack{Num.\\Est.} & \shortstack{Visual\\Ground.} & \shortstack{Spatial Rel.\\Assess.} & \shortstack{Depth\\Order.} & \shortstack{Instr. Func.\\Ident.} & \shortstack{Behavior\\Class.} \\
\midrule
Nautilus-7B~\citep{xu2025nautilus} & 11.00 & 33.00 & 7.00 & 24.00 & 60.00 & 36.00 \\
OceanGPT-7B~\citep{bi2023oceangpt} & 20.00 & 20.00 & 2.00 & 0.00 & 28.00 & 1.00 \\
Qwen3-VL-8B~\citep{bai2025qwen3vl} & \secondcell 34.00 & \secondcell 84.00 & \secondcell 14.00 & \secondcell 29.00 & \secondcell 66.00 & \secondcell 63.00 \\
USI-Brain (ours) & \bestcell \textbf{46.00} & \bestcell \textbf{93.00} & \bestcell \textbf{29.00} & \bestcell \textbf{31.00} & \bestcell \textbf{77.00} & \bestcell \textbf{69.00} \\
\bottomrule
\end{tabular*}
\end{adjustbox}\par

\vspace{0.4em}

\begin{adjustbox}{max width=\linewidth}
\begin{tabular*}{\linewidth}{@{\extracolsep{\fill}}lcccccc@{}}
\toprule
Method & \shortstack{Traits\\Extract.} & \shortstack{Threat Status\\Determ.} & \shortstack{Pollutant\\Localiz.} & \shortstack{Disaster\\Diagn.} & \shortstack{Species\\Compr.} & \shortstack{Marine Tech.\\Underst.} \\
\midrule
Nautilus-7B~\citep{xu2025nautilus} & 77.00 & 15.00 & 33.00 & \bestcell \textbf{94.00} & 27.70 & 60.00 \\
OceanGPT-7B~\citep{bi2023oceangpt} & 33.00 & 0.00 & 31.00 & 85.00 & 18.00 & 28.00 \\
Qwen3-VL-8B~\citep{bai2025qwen3vl} & \secondcell 87.00 & \secondcell 32.00 & \bestcell \textbf{40.00} & \secondcell 93.00 & \bestcell \textbf{51.80} & \secondcell 71.00 \\
USI-Brain (ours) & \bestcell \textbf{89.00} & \bestcell \textbf{51.00} & \secondcell 37.00 & \bestcell \textbf{94.00} & \secondcell 43.83 & \bestcell \textbf{72.00} \\
\bottomrule
\end{tabular*}
\end{adjustbox}\par
\end{minipage}\par\medskip

USI-Brain achieves the highest scores in nine categories and ties with Nautilus-7B on disaster diagnosis. Relative to the strongest baseline for each category, the gains include 15 points in spatial-relation assessment, 12 in numerical estimation, 11 in instrument-function identification, and 19 in threat-status determination. Qwen3-VL-8B scores higher on pollutant localization and species comprehension in this comparison.

\Needspace{5\baselineskip}
\section{Data and Component Ablations}
\label{app:data-component-ablations}

We vary the capability groups included in training and evaluate every configuration on all 15 USI-Bench tasks. Table~\ref{tab:data-component-ablations} lists the training groups and enabled components. Base is Qwen3-VL-8B-Instruct; B1--B5 follow Section~\ref{sec:benchmark}. The UPR comparison uses matched training QA and settings, while the source-removal and group-subset comparisons vary the included supervision.

\begingroup
\centering
\small
\setlength{\tabcolsep}{4.0pt}
\renewcommand{\arraystretch}{1.10}
\captionof{table}{Data and component ablations on USI-Bench. Base reports absolute scores; every other row reports signed gains over Base. Overall equally weights B1--B5. $\checkmark$/$\times$ indicate enabled/disabled components; --- denotes not applicable. Boldface marks the largest gain. Darker/lighter blue marks the best/second-best distinct gains among the ablations, excluding Base. Positive/negative differences are red/green.}
\label{tab:data-component-ablations}
\begin{adjustbox}{max width=\linewidth}
\begin{tabular*}{\linewidth}{@{\extracolsep{\fill}}lccccccccc@{}}
\toprule
\multirow{2}{*}{Training groups} & \multicolumn{3}{c}{Components} & \multicolumn{6}{c}{Base scores / signed gains} \\
\cmidrule(lr){2-4} \cmidrule(lr){5-10}
 & HUE & GeoQA & UPR & B1 & B2 & B3 & B4 & B5 & Overall \\
\midrule
Base & --- & --- & $\times$ & 55.96 & 67.85 & 61.45 & 63.85 & 50.60 & 59.94 \\
\midrule
B1 & $\times$ & $\times$ & $\times$ & \posdelta{+3.10} & \posdelta{+0.20} & \posdelta{+0.50} & \negdelta{$-0.20$} & \negdelta{$-0.10$} & \posdelta{+0.70} \\
B1+B2 & $\times$ & $\times$ & $\times$ & \posdelta{+3.00} & \posdelta{+1.60} & \posdelta{+0.90} & \negdelta{$-0.10$} & \negdelta{$-0.20$} & \posdelta{+1.04} \\
B1+B3 & $\times$ & $\checkmark$ & $\times$ & \posdelta{+3.20} & \posdelta{+0.80} & \posdelta{+4.60} & \posdelta{+0.10} & \negdelta{$-0.10$} & \posdelta{+1.72} \\
B2+B3 & $\times$ & $\checkmark$ & $\times$ & \posdelta{+0.10} & \posdelta{+1.90} & \posdelta{+4.30} & \negdelta{$-0.20$} & \negdelta{$-0.30$} & \posdelta{+1.16} \\
B1+B2+B3 & $\times$ & $\checkmark$ & $\times$ & \posdelta{+3.10} & \posdelta{+1.90} & \posdelta{+5.10} & \posdelta{+0.20} & \negdelta{$-0.20$} & \posdelta{+2.02} \\
B1--B4 & $\times$ & $\checkmark$ & $\times$ & \posdelta{+3.30} & \posdelta{+1.80} & \posdelta{+5.20} & \posdelta{+3.00} & \negdelta{$-0.10$} & \posdelta{+2.64} \\
B1--B5 & $\times$ & $\checkmark$ & $\times$ & \posdelta{+4.88} & \posdelta{+1.75} & \posdelta{+5.28} & \posdelta{+4.00} & \posdelta{+1.10} & \posdelta{+3.40} \\
\midrule
B1+B2+B4+B5 & $\checkmark$ & $\times$ & $\times$ & \posdelta{+7.18} & \negdelta{$-1.00$} & \posdelta{+2.58} & \posdelta{+5.20} & \posdelta{+2.05} & \posdelta{+3.20} \\
B1--B5 & $\checkmark$ & $\checkmark$ & $\times$ & \secondcell \posdelta{+7.58} & \secondcell \posdelta{+2.35} & \secondcell \posdelta{+6.65} & \secondcell \posdelta{+6.15} & \secondcell \posdelta{+2.45} & \secondcell \posdelta{+5.04} \\
B1--B5 (Full) & $\checkmark$ & $\checkmark$ & $\checkmark$ & \bestcell \textbf{\posdelta{+8.54}} & \bestcell \textbf{\posdelta{+3.35}} & \bestcell \textbf{\posdelta{+7.18}} & \bestcell \textbf{\posdelta{+6.55}} & \bestcell \textbf{\posdelta{+11.60}} & \bestcell \textbf{\posdelta{+7.44}} \\
\bottomrule
\end{tabular*}
\end{adjustbox}\par
\endgroup
\par\medskip

\subsection{Aggregation and Reporting}
Within each capability group, we average the task scores reported on the 100-point scale. B1 contains Identity, Grounding, Counting, Detection, and Captioning; B2 contains Relation and Distance; B3 contains Depth, Viewpoint, Ordering, and Fg./Bg.; B4 contains Environment and Structure; B5 contains Scene and Ecology. The group sizes are therefore 5, 2, 4, 2, and 2.

\Needspace{7\baselineskip}
For model $m$, let $\mathcal{T}_b$ be the task set of group $b$ and $s_{m,t}$ its score on task $t$. We define
\begin{equation}
\begin{aligned}
S_{m,b} &= \frac{1}{|\mathcal{T}_b|}\sum_{t\in\mathcal{T}_b}s_{m,t},
\qquad \Delta_{m,b}=S_{m,b}-S_{\mathrm{Base},b},\\
S_{m,\mathrm{Overall}} &= \frac{1}{5}\sum_{b=1}^{5}S_{m,b}.
\end{aligned}
\label{eq:ablation-group-aggregate}
\end{equation}
The Overall gain equally weights the five group gains, whereas the main-text average equally weights the 15 tasks. This yields +5.04/+7.44 for USI-Brain/Full, compared with +5.76/+7.63 under task weighting. Means and differences are computed before rounding to two decimals; subtracting rounded entries can differ by 0.01.

\subsection{Configuration Correspondence}
B1--B5 with HUE and GeoQA but without UPR is USI-Brain; enabling UPR gives Full. B1--B5 without HUE maps to \emph{w/o HUE}; B1+B2+B4+B5 with HUE and without GeoQA maps to \emph{w/o GeoQA}. Both removals disable UPR. These four rows reproduce Table~\ref{tab:main-results}'s group means. Full and USI-Brain use the same 4.27M QA and training settings; all variants use raw evaluation images.

\subsection{Effects of Training Coverage and Components}
\textbf{Training-group coverage.} Expanding training coverage produces gains in the corresponding capability groups: adding B3 to B1 training increases the B3 gain from +0.50 to +4.60 points, while adding B4 to B1+B2+B3 increases the B4 gain from +0.20 to +3.00 points. 

\textbf{Supervision sources.} With all five groups included and UPR disabled, HUE raises B1 and B4 scores by 2.70 and 2.15 points. Removing GeoQA from USI-Brain lowers the B3 gain from +6.65 to +2.58 and the B2 gain from +2.35 to $-1.00$. 

\textbf{Feature rectification.} Adding UPR to USI-Brain improves B5 by a further 9.15 points, with smaller gains of 0.40--1.00 points across B1--B4. The B5 change averages the 18.0-point Scene gain and the 0.3-point Ecology gain.

\Needspace{5\baselineskip}
\section{Inference Efficiency}
\label{app:inference-efficiency}

Table~\ref{tab:inference-efficiency} summarizes inference on a single GPU with BF16 and batch size 1. A800 runs use Transformers with FlashAttention 2; RTX 3090 (24 GB) runs use Transformers 4.57.1 with SDPA. USI-Brain disables UPR, while USI-Brain + UPR includes the prior encoder and feature-rectification path. All configurations use raw input images. The input sequence lengths and attention backends follow the configuration reported for each GPU.

\par\medskip\noindent\begin{minipage}{\textwidth}
\centering
\small
\setlength{\tabcolsep}{4.0pt}
\renewcommand{\arraystretch}{1.15}
\captionof{table}{Single-GPU inference efficiency. In/Out denotes input/generated token counts; E2E denotes end-to-end latency. USI-Brain latencies are medians, and its RTX 3090 token counts and decoding speeds are also medians. Memory is PyTorch peak GPU memory.}
\label{tab:inference-efficiency}
\begin{adjustbox}{max width=\linewidth}
\begin{tabular*}{\linewidth}{@{\extracolsep{\fill}}llccccc@{}}
\toprule
GPU & Model & Images & \shortstack{In/Out\\tokens} & \shortstack{Decode\\(tokens/s)} & \shortstack{E2E\\(s)} & \shortstack{Peak memory\\(GiB)} \\
\midrule
A800 & USI-Brain & 1 & 3117 / 21.0 & 27.50 & 1.35 & 17.24 \\
A800 & USI-Brain & 2 & 6210 / 25.0 & 27.70 & 2.08 & 18.14 \\
\midrule
RTX 3090 & USI-Brain & 1 & 793 / 18.5 & 35.86 & 0.84 & 16.57 \\
RTX 3090 & USI-Brain & 2 & 1563 / 17.5 & 33.26 & 1.12 & 16.79 \\
RTX 3090 & USI-Brain + UPR & 1 & 1561 / 20.0 & 34.50 & 1.63 & 21.80 \\
RTX 3090 & USI-Brain + UPR & 2 & 3099 / 26.5 & 30.90 & 2.95 & 23.60 \\
\bottomrule
\end{tabular*}
\end{adjustbox}\par
\end{minipage}\par\medskip

On A800, median time to first token, including preprocessing, is 0.63 s for one image and 1.21 s for two views. The corresponding E2E P95 values are 1.37 s and 2.08 s. On RTX 3090, USI-Brain answers single-image queries in a median of 0.84 s using 16.57 GiB. USI-Brain + UPR records 1.63 s latency and 21.80 GiB peak memory for a single image, and 2.95 s and 23.60 GiB for two images.


\end{document}